 \documentclass[pmlr,twocolumn,10pt]{jmlr} 

\usepackage{booktabs}
\usepackage{multirow}
\usepackage{listings}
\usepackage{algorithm}
\usepackage[noend]{algpseudocode}
\usepackage[belowskip=-1pt,aboveskip=5pt]{caption}

\usepackage[load-configurations=version-1]{siunitx} 

\newcommand{\equal}[1]{{\hypersetup{linkcolor=black}\thanks{#1}}}

\theorembodyfont{\upshape}
\theoremheaderfont{\scshape}
\theorempostheader{:}
\theoremsep{\newline}

\jmlrvolume{LEAVE UNSET}
\jmlryear{2021}
\jmlrsubmitted{LEAVE UNSET}
\jmlrpublished{LEAVE UNSET}
\jmlrworkshop{Machine Learning for Health (ML4H) 2021} 

\title{Adding more data does not always help: A study in medical conversation summarization with PEGASUS}

\author{
  Varun Nair\equal{Duke University. Work done while author was a research intern at Curai.}  \Email{vn40@duke.edu}\\
  Namit Katariya\Email{namit@curai.com}\\
 Xavier Amatriain \Email{xavier@curai.com} \\
  Ilya Valmianski \Email{ilya@curai.com}\\
  Anitha Kannan  \Email{anitha@curai.com} \\
  \addr Curai 
}

\begin{document}

\maketitle
\begin{abstract}

Medical conversation summarization is integral in capturing information gathered during interactions between patients and physicians. Summarized conversations are used to facilitate patient hand-offs between physicians, and as part of providing care in the future. Summaries, however, can be time-consuming to produce and require domain expertise. Modern pre-trained NLP models such as PEGASUS have emerged as capable alternatives to human summarization, reaching state-of-the-art performance on many summarization benchmarks. However, many downstream tasks still require at least moderately sized datasets to achieve satisfactory performance. In this work we  (1) explore the effect of dataset size on transfer learning medical conversation summarization using PEGASUS and (2) evaluate various iterative labeling strategies in the low-data regime, following their success in the classification setting. We find that model performance saturates with increase in dataset size and that the various active-learning strategies evaluated all show equivalent performance consistent with simple dataset size increase. We also find that naive iterative pseudo-labeling is on-par or slightly worse than no pseudo-labeling. Our work sheds light on the successes and challenges of translating low-data regime techniques in classification to medical conversation summarization and helps guides future work in this space. Relevant code available at \url{https://github.com/curai/curai-research/tree/main/medical-summarization-ML4H-2021}.

\end{abstract}

\section{Introduction}

Medical conversation summarization can help medical providers to keep a record of patient encounters and also provide the necessary context of a patient's medical history during patient hand-offs between providers. However, creating these summaries presents a significant clerical load on medical providers, which can lead to burnout \citep{shanafelt2016}.

To tackle this problem, we evaluate the use of transfer learning with abstractive summarization models, such as PEGASUS \citep{zhang2019}, to fine tune on medical conversation summarization. Availability of pre-trained models has led to tremendous progress in multiple other domains \citep{yadav2021,dai2021,xu2021}, with fine-tuning being the main strategy to derive task specificity. Notably, many tasks still require at least moderately sized datasets to capture all task-specific regularities. In the medical domain especially, obtaining labels is resource-intensive and dependent on domain expertise. 

Much of the recent work in label-scarce and low-data regime domains has focused on either effectively collecting informative labeled data or leveraging unlabeled data \citep{mindermann2021, du2020, chen2020}. The goal is to label enough informative samples such that the overall need for expert-labels can be alleviated while still achieving satisfactory performance. This can be done via active learning, in which an acquisition function is used to iteratively inform the labeling process and select the most ``useful" points to label. An alternate approach that has regained popularity is self-training or pseudo-labeling, in which the model's predictions are used as ground truth in subsequent re-trained iterations of a model. In particular, \cite{du2020} show that self-training with pseudo-labeling can improve performance on text classification benchmarks without the need for in-domain unlabeled data.


Given the context of the success of active learning and pseudo-labeling for low-data regime classification tasks, we are interested in the following: 

\begin{itemize}
    \item Will these strategies translate  from the classification setting to the generative task of medical conversation summarization?
    \item Is there (empirical) consensus on how we can most effectively utilize our label budget to maximize summarization performance with these strategies?
\end{itemize}

In this work, we explore these questions by applying batched active-learning to the task of medical conversation summarization in the low-data regime, implemented by leveraging model confidence on unlabeled samples as signal for expert-labeling and pseudo-labeling. Following our exploration, the contributions of this work are as follows:

\begin{itemize}
    \item We find that among the strategies for selecting labels to be expert-labeled, no one strategy emerges as the best. Furthermore, performance on our conversation-summary pair dataset saturates as the number of expert labels increase, while still falling below the theoretical maximum metric values (see \S~\ref{subsec:metrics}).
    \item We find that naive pseudo-labeling is on-par or slightly worse than using no pseudo-labeling, often providing the biggest increase following the first iteration of self-training.
\end{itemize}

\section{Experimental Setup}
\label{sec:experimental-setup}


\subsection{Approaches Considered}
Medical conversation summarization suffers  not only from lack of annotated data but also from a high cost of annotating additional examples \citep{chintagunta-etal-2021-medically, joshi2020}. We investigate\footnote{Code at \url{https://github.com/curai/curai-research/tree/main/medical-summarization-ML4H-2021}} how to collect more labeled examples to improve PEGASUS \citep{zhang2019}, a state-of-the-art abstractive summarization model, on the task of medical conversation summarization while starting with only a small amount of human labeled data.

Algorithm~\autoref{alg:training} describes our overall strategy to add labeled data following each iteration $i$ of self-training. We experiment with two ways to add labeled points to our training set.
\begin{enumerate}
    \item \textbf{Expert/Human labeling (\textsf{HL})}: Have \textit{medical-expert} summarized conversations added to the labeled set, given that PEGASUS \textit{is not likely to do well} on these samples.
    \item \textbf{Pseudo-labeling (\textsf{PL})}: Have \textit{model-generated} summaries added to the labeled set, given that PEGASUS is \textit{likely to do well} on them.
\end{enumerate}
We allow the strategies for pseudo-labeling (\textsf{PL}) and expert-labeling (\textsf{HL}) to return empty sets. This allows us to experiment with pseudo-labeling and expert-labeling strategies in isolation as well as in tandem. For both pseudo-labeling and human labeling, we experiment with a budget of zero and 1\% of the size of $U_0$.

\begin{algorithm}[H]
  \caption{Iterative labeling strategy}
  \label{alg:training}
  \begin{algorithmic}[1]
    \Require
    \Statex labeled set $\mathcal{L}_0$ 
    \Statex unlabeled set $U_0$
    \Statex pretrained summarization model PEGASUS $M_0$
    \Statex labeling budgets $b_P$ for pseudo-labeling
    \Statex labeling budgets $b_E$ for expert-labeling,
    \Statex total number of iterations for self-training $n$
    \Statex
      \For{$i \gets 1, \cdots, n$}
        \State $M_i \gets$ \text{Train $M_0$ on $\mathcal{L}_{i-1}$}
        \State $s_i \gets \text{log likelihood of }M_i \text{ on } U_{i-1}$
        \State $P_i \gets \text{Pseudo-label $\textsf{PL}(s_i, U_{i-1}, b_P)$} \text{ using } M_i$
        \State $E_i \gets \text{Expert-label $\textsf{HL}(s_i, U_{i-1}, b_E)$}$
        \State $\mathcal{L}_i \gets \mathcal{L}_{i-1} + P_i + E_i$
        \State $U_i \gets U_{i-1} - P_i - E_i$
      \EndFor
      \State \Return $M_{n}$
  \end{algorithmic}
\end{algorithm}

We derive our confidence score of how well PEGASUS can generate a summary for a given data point by using the log-likelihood of the generated summary as a proxy. The log-likelihood of class predictions has been used in prior work for classification tasks in a similar manner \citep{xie2019,sohn2020}.

\subsection{Experiment Configurations}
\label{subsec:expconfig}
We examine performance characteristics of PEGASUS fine tuned on datasets acquired with a large number of experimental configurations. For each combination described in  Table~\ref{tab:experiments}, we ran three additional iterations when either human or pseudo-labeling strategy was not \verb|None|, for a total of 264 experiments (some iteration 0 experiments are equivalent and thus were not performed, since some conditions only affect later iterations of self-training). For experiments differentiated only by the dropout value, 0.1 (default) and 0.5 (large regularization), the reported result is the best of the two.

\begin{table}[t]
    \centering
    \begin{tabular}{ll}
    \toprule
    \textbf{Condition} & \textbf{Variants} \\
    \midrule
        Dropout &  0.1 \\ & 0.5 \\
        \midrule
        Pseudo-labeling strategy (\textsf{PL}) & None \\ 
                                 & Top 1\% \\
                                 \midrule
        Human-labeling strategy (\textsf{HL})  & None \\ 
                                 & Bottom 1\% \\  
                                 & Middle 1\% \\  
                                 & Random 1\% \\ 
                                 \midrule
        Starting number of samples ($|\mathcal{L}_0|$)  & 100 \\ 
                                 & 250 \\  
                                 & 500 \\  
                                 & 750 \\  
                                 & 1000 \\  
                                 & 1250  \\
    \bottomrule
    \end{tabular}
    \caption{Overview of experimental conditions as detailed in \autoref{subsec:expconfig}.}
    \label{tab:experiments}
\end{table}

\subsection{Training Details and Metrics}
\label{sec:training-details}
\label{subsec:metrics}
\begin{figure*}[!tbp]
    \centering
    \includegraphics[width=1\linewidth]{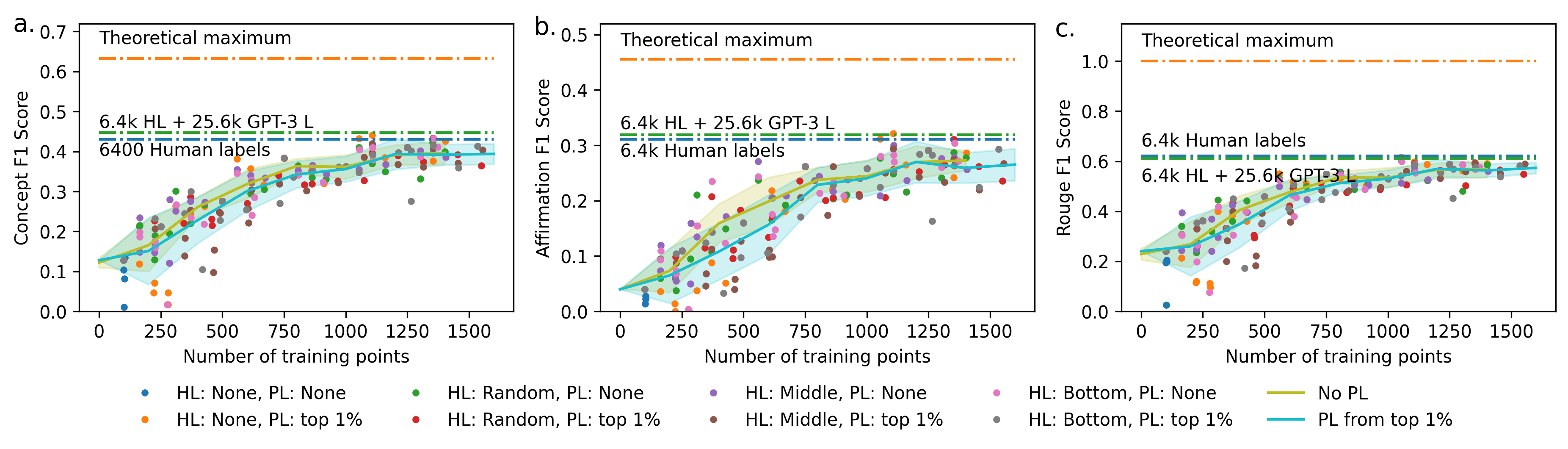}
    \caption{(a) Concept, (b) Affirmation, and (c) Rouge F1 scores as a function of number of training points and sampling strategy. Shown are two trend-lines and their standard deviations for models trained with and without pseudo-labeled data (``PL from top 1\%'' and ``No PL'', respectively). The horizontal lines represent performance for: theoretical maximum, dataset of 6400 human labels, and combined dataset of 6400 human labels and 25600 GPT-3 derived labels \citep{chintagunta-etal-2021-medically}. Note: this paper does {\bf not} study the role of GPT-3 as a function of training set size, which we leave as future work.}
    \label{fig:saturation}
\end{figure*}

We replicate the experimental settings with small modifications\footnote{See Appendix \autoref{sec:appendix_training_details} for further training details.} for PEGASUS from \cite{chintagunta-etal-2021-medically}  by importing the pre-trained PEGASUS model on the CNN/DailyMail summarization dataset from HuggingFace.\footnote{\url{https://huggingface.co/transformers/model\_doc/pegasus.html}} We apply these experimental settings uniformly across all iterations of self-training as well as all baselines plotted on Figure \ref{fig:saturation}.

Performance of our model was measured using the same metrics defined in \citep{joshi2020, chintagunta-etal-2021-medically}  viz. concept F1, affirmation F1 and rouge-L F1. We also computed the theoretical maximum of these metrics on our test set. The maximum value for concept F1 and affirmation F1 is computed by assuming that the predicted summary is the same as the ground truth label, i.e. we predict both the concepts and their affirmations correctly for each conversation.\footnote{We provide further explanation of these metrics and the theoretical maximum in Appendix \autoref{sec:appendix_metrics}.}

\section{Results}
\label{sec:results}

We use a human expert labeled dataset from \cite{chintagunta-etal-2021-medically} to study the effect of dataset size and iterative labeling strategies (Table~\ref{tab:experiments}). We present two main findings: (1) performance saturation with dataset size (2) effect of iterative labeling strategies in the low data regime. 

\subsection{Performance saturation with dataset size}

Our main finding is the saturation of model performance as the number of dataset samples increase. An overview of the results can be seen in \figureref{fig:saturation}. 
\begin{enumerate}
    \item Setting $\textsf{PL} = \textsf{HL} = \textsf{None}$ and $n=1$, we find model performance saturates quickly. Specifically, we find the model with $|\mathcal{L}_0| = 1250$ achieves similar performance to $|\mathcal{L}_0| = 6400$, with concept F1 scores of 0.41 vs 0.43 respectively.
    \item We also find the $(n = 2, |\mathcal{L}_0| = 1000, |\mathcal{L}_2| = 1159, \textsf{PL} = \textsf{Top 1\%}, \textsf{HL} = \textsf{None})$ model beats the $(n=1, |\mathcal{L}_0| = 6400, \textsf{PL} = \textsf{HL} = \textsf{None})$ model, with concept F1 scores 0.44 vs 0.43 respectively.
\end{enumerate}

Furthermore, we compare our results with \cite{chintagunta-etal-2021-medically}, with whom we share the test-set. We find that our best low-data models achieve similar results to the baselines with significantly more human-labeled points and to the baseline with both human-labeled and GPT-3 labeled points. Notably, \cite{chintagunta-etal-2021-medically} saw continued improvement with additional data while we do not, which we suspect to be due to insufficient optimization of training hyper-parameters (see \ref{sec:training-details} for a comparison of hyper-parameters). Notably, however, we did not study the utility of GPT-3 generated labels in a low-data regime and leave this to future work.

\vspace{1em}
We also observe the saturation behavior by comparing our achieved metrics with the maximum theoretically achievable values. We find that performance is often significantly below maximum values (e.g. concept F1 of 0.44 vs 0.63, affirmation F1 of 0.31 vs 0.45, and rouge F1 of 0.62 vs 1.0 respectively). We suspect that the early saturation below the maximum theoretical values reflect the limits of PEGASUS on the medical conversation summarization task. An interesting future work would include evaluating the saturation behavior of other pre-trained abstractive summarization models.

\subsection{Effect of iterative labeling strategies in low data regime}

In the low data regime we examined several iterative labeling strategies consisting of adding additional human and pseudo-labeled examples (see Table~\ref{tab:experiments}).

From \figureref{fig:saturation}, we find that for a given number of training points, $\textsf{PL} = \textsf{None}$ experiments have larger mean value than $\textsf{PL} = \textsf{Top 1\%}$ experiments for every metric for most values of  $|\mathcal{L}_0|$ and $i$. However, these results are only statistically significant for affirmation F1 scores with $\approx 500$ training points.

Overall, the performance improved with additional human labeled data consistently with the number of points added. A more detailed view of the effect of human-labeling can be seen from \figureref{fig:sampling_strat}. For our studies of up to three additional iterations, we did not find that the performance is impacted by the choice of human-labeling strategies $\textsf{HL}$, as \textsf{Top 1\%}, \textsf{Bottom 1\%} and \textsf{Random 1\%} produced similar results. We did find that in the low data regime, human labels are necessary. In \figureref{fig:sampling_strat} looking at $\textsf{HL} = \textsf{None}$ strategy, we find that pseudo-labels are ineffective at improving test performance when $|\mathcal{L}_0| = 100$.

\section{Discussion}

We extended active-learning and pseudo-labeling techniques to the generative setting following recent success in the classification setting. We found that for PEGASUS, a SOTA summarization model, there is a rapid performance saturation at ~1000 examples when fine tuned for medical conversation summarization. Beyond a small number of examples, providing additional labels did not improve performance. Even with few labels, we found that the choice of active labeling strategy had little effect on the performance, rather, the number of samples ended up being paramount. 

Compared to previous results, we also found performance saturation to occur at a lower sample count, due to hyper-parameter optimization done for fine-tuning on a smaller dataset. This provides an important lesson on ensuring hyper-parameters are well tuned before requesting more data.

Through the lens of the bias-variance trade off, we believe PEGASUS represents a high bias model and this potentially allows it to generalize across domains. However, it cannot fully adapt to hyper-specialized regularities, as we struggled to over-fit to the train set. This explains why PEGASUS quickly saturates on medical dialogue data and why all sampling strategies are somewhat equivalent, since low confidence examples contain medical jargon not present during PEGASUS pre-training and tokenizer initialization. Understanding how such text generation models fine tune to fields with significant field-specific jargon presents an important area for future exploration.




\bibliography{reference}

\begin{thebibliography}{14}
\providecommand{\natexlab}[1]{#1}
\providecommand{\url}[1]{\texttt{#1}}
\expandafter\ifx\csname urlstyle\endcsname\relax
  \providecommand{\doi}[1]{doi: #1}\else
  \providecommand{\doi}{doi: \begingroup \urlstyle{rm}\Url}\fi

\bibitem[Chen et~al.(2020)Chen, Yang, and Yang]{chen2020}
Jiaao Chen, Zichao Yang, and Diyi Yang.
\newblock Mixtext: Linguistically-informed interpolation of hidden space for
  semi-supervised text classification.
\newblock \emph{CoRR}, abs/2004.12239, 2020.
\newblock URL \url{https://arxiv.org/abs/2004.12239}.

\bibitem[Chintagunta et~al.(2021)Chintagunta, Katariya, Amatriain, and
  Kannan]{chintagunta-etal-2021-medically}
Bharath Chintagunta, Namit Katariya, Xavier Amatriain, and Anitha Kannan.
\newblock Medically aware {GPT}-3 as a data generator for medical dialogue
  summarization.
\newblock In \emph{Proceedings of the Second Workshop on Natural Language
  Processing for Medical Conversations}, pages 66--76, Online, June 2021.
  Association for Computational Linguistics.
\newblock \doi{10.18653/v1/2021.nlpmc-1.9}.
\newblock URL \url{https://aclanthology.org/2021.nlpmc-1.9}.

\bibitem[Dai et~al.(2021)Dai, Wang, Lyu, and Zhu]{dai2021}
Songtai Dai, Quan Wang, Yajuan Lyu, and Yong Zhu.
\newblock Bdkg at mediqa 2021: System report for the radiology report
  summarization task.
\newblock In \emph{BIONLP}, 2021.

\bibitem[Du et~al.(2020)Du, Grave, Gunel, Chaudhary, Celebi, Auli, Stoyanov,
  and Conneau]{du2020}
Jingfei Du, Edouard Grave, Beliz Gunel, Vishrav Chaudhary, Onur Celebi, Michael
  Auli, Ves Stoyanov, and Alexis Conneau.
\newblock Self-training improves pre-training for natural language
  understanding.
\newblock \emph{CoRR}, abs/2010.02194, 2020.
\newblock URL \url{https://arxiv.org/abs/2010.02194}.

\bibitem[Harkema et~al.(2009)Harkema, Dowling, Thornblade, and
  Chapman]{negex09}
Henk Harkema, John~N. Dowling, Tyler Thornblade, and Wendy~W. Chapman.
\newblock Context: An algorithm for determining negation, experiencer, and
  temporal status from clinical reports.
\newblock \emph{Journal of Biomedical Informatics}, 42\penalty0 (5):\penalty0
  839 -- 851, 2009.
\newblock Biomedical Natural Language Processing.

\bibitem[Joshi et~al.(2020)Joshi, Katariya, Amatriain, and Kannan]{joshi2020}
Anirudh Joshi, Namit Katariya, Xavier Amatriain, and Anitha Kannan.
\newblock Dr. summarize: Global summarization of medical dialogue by exploiting
  local structures.
\newblock In \emph{Empirical Methods in Natural Language Processing (EMNLP)
  Findings}, 2020.

\bibitem[Lin(2004)]{lin-2004-rouge}
Chin-Yew Lin.
\newblock {ROUGE}: A package for automatic evaluation of summaries.
\newblock In \emph{Text Summarization Branches Out}, pages 74--81, Barcelona,
  Spain, July 2004. Association for Computational Linguistics.
\newblock URL \url{https://www.aclweb.org/anthology/W04-1013}.

\bibitem[Mindermann et~al.(2021)Mindermann, Razzak, Xu, Kirsch, Sharma,
  Morisot, Gomez, Farquhar, Brauner, and Gal]{mindermann2021}
Sören Mindermann, Muhammed Razzak, Winnie Xu, Andreas Kirsch, Mrinank Sharma,
  Adrien Morisot, Aidan~N. Gomez, Sebastian Farquhar, Jan Brauner, and Yarin
  Gal.
\newblock Prioritized training on points that are learnable, worth learning,
  and not yet learned.
\newblock 2021.

\bibitem[Shanafelt et~al.(2016)Shanafelt, Dyrbye, Sinsky, Hasan, Satele, Sloan,
  and West]{shanafelt2016}
Tait~D. Shanafelt, Lotte~N. Dyrbye, Christine Sinsky, Omar Hasan, Daniel
  Satele, Jeff Sloan, and Colin~P. West.
\newblock Relationship between clerical burden and characteristics of the
  electronic environment with physician burnout and professional satisfaction.
\newblock \emph{Mayo Clinic Proceedings}, 91\penalty0 (7):\penalty0 836--848,
  2016.
\newblock ISSN 0025-6196.
\newblock \doi{https://doi.org/10.1016/j.mayocp.2016.05.007}.
\newblock URL
  \url{https://www.sciencedirect.com/science/article/pii/S0025619616302154}.

\bibitem[Sohn et~al.(2020)Sohn, Berthelot, Li, Zhang, Carlini, Cubuk, Kurakin,
  Zhang, and Raffel]{sohn2020}
Kihyuk Sohn, David Berthelot, Chun{-}Liang Li, Zizhao Zhang, Nicholas Carlini,
  Ekin~D. Cubuk, Alex Kurakin, Han Zhang, and Colin Raffel.
\newblock Fixmatch: Simplifying semi-supervised learning with consistency and
  confidence.
\newblock \emph{CoRR}, abs/2001.07685, 2020.
\newblock URL \url{https://arxiv.org/abs/2001.07685}.

\bibitem[Xie et~al.(2019)Xie, Dai, Hovy, Luong, and Le]{xie2019}
Qizhe Xie, Zihang Dai, Eduard~H. Hovy, Minh{-}Thang Luong, and Quoc~V. Le.
\newblock Unsupervised data augmentation.
\newblock \emph{CoRR}, abs/1904.12848, 2019.
\newblock URL \url{http://arxiv.org/abs/1904.12848}.

\bibitem[Xu et~al.(2021)Xu, Zhang, Hong, Cai, and Sung]{xu2021}
Liwen Xu, Yan Zhang, Lei Hong, Yi~Cai, and Szui Sung.
\newblock Chichealth @ mediqa 2021: Exploring the limits of pre-trained seq2seq
  models for medical summarization.
\newblock In \emph{BIONLP}, 2021.

\bibitem[Yadav et~al.(2021)Yadav, Gupta, Abacha, and
  Demner{-}Fushman]{yadav2021}
Shweta Yadav, Deepak Gupta, Asma~Ben Abacha, and Dina Demner{-}Fushman.
\newblock Question-aware transformer models for consumer health question
  summarization.
\newblock \emph{CoRR}, abs/2106.00219, 2021.
\newblock URL \url{https://arxiv.org/abs/2106.00219}.

\bibitem[Zhang et~al.(2019)Zhang, Zhao, Saleh, and Liu]{zhang2019}
Jingqing Zhang, Yao Zhao, Mohammad Saleh, and Peter~J. Liu.
\newblock {PEGASUS:} pre-training with extracted gap-sentences for abstractive
  summarization.
\newblock \emph{CoRR}, abs/1912.08777, 2019.
\newblock URL \url{http://arxiv.org/abs/1912.08777}.

\end{thebibliography}

\counterwithin{figure}{section}
\counterwithin{table}{section}
\clearpage
\appendix
\section{Training Details}
\label{sec:appendix_training_details}
We follow most experimental settings from \cite{chintagunta-etal-2021-medically}, but make modifications to the length of training to 6 epochs and increase the effective batch size to 128 for most experiments.\footnote{Experiments with fewer than 128 samples have an adjusted effective batch size that is 2-4x smaller to allow for adequate training.} These modifications were made for better optimization in the lower-data regime, compared to the 6400+ samples used in experiments from \cite{chintagunta-etal-2021-medically}. We note again that these experimental settings are applied uniformly across all iterations of self-training as well as all baselines plotted on Figure \ref{fig:saturation}.

\section{Metrics}
\label{sec:appendix_metrics}
We measure model performance on standard metrics of ROUGE \citep{lin-2004-rouge} \footnote{\raggedright We use the following package with default configuration: \url{https://github.com/google-research/google-research/tree/master/rouge}} as well as measure a model's effectiveness in capturing the  medical concepts that are of importance, and their negations \cite{joshi2020}.

\noindent {\bf Medical Concept Coverage}: The concept coverage set of metrics captures the coverage of medical terms in the model's output summary with respect to the ground truth. In particular, let $\mathcal{C}$  be the set of medical concepts in the reference summary and $\hat{\mathcal{C}} $ be the set of concepts in the summary output by the model. Then:
$$\textrm{Concept recall} = \frac{\sum_{n=1}^{N}  |\hat{\mathcal{C}}^{(n)}  \cap  \mathcal{C}^{(n)}| }  {\sum_{n=1}^{N}|\mathcal{C}^{(n)}|} $$$$\textrm{Concept precision} = \frac{\sum_{n=1}^{N}  |\hat{\mathcal{C}}^{(n)}  \cap  \mathcal{C}^{(n)}| }  {\sum_{n=1}^{N}|\hat{\mathcal{C}}^{(n)}|}$$. 

We use these to compute a concept F1\footnote{Note if there are no concepts detected in the snippet and summary by the entity extractor, then a conservative F1 score of 0 is given for that example.\label{footnote_f1}} and use an in-house medical entity extractor to extract medical concepts in the summary. Medical concepts in the decoded summary that weren't present in the original conversation would be false positives and vice versa for false negatives.
 
\noindent {\bf Affirmation Correctness}:  To measure the effectiveness of the model to identify the affirmative (alternatively, negated) status of medical concepts,
we use Negex \citep{negex09} to determine negated concepts. Of the concepts present in the decoded summary, we evaluate precision and recall on whether the affirmations and negations were accurate for the decoded concepts and compute an affirmation F1.$^{\ref{footnote_f1}}$

\noindent {\bf Theoretical Maximum Derivation}: The maximum score of the concept and affirmation metrics is not 1.0 because, as in \cite{chintagunta-etal-2021-medically}, a conservative concept as well as affirmation F1 score of 0 is given for an example if there are no concepts detected in the conversation. As our affirmation tagger is not perfect, it may also ignore/miss certain concepts leading to the affirmation F1 limit not being identical to the concept F1 limit.

\section{Detailed view into sampling strategy experiments}
\label{sec:appendix}

\begin{figure}[!h]
    \centering
    \includegraphics[width=1\linewidth]{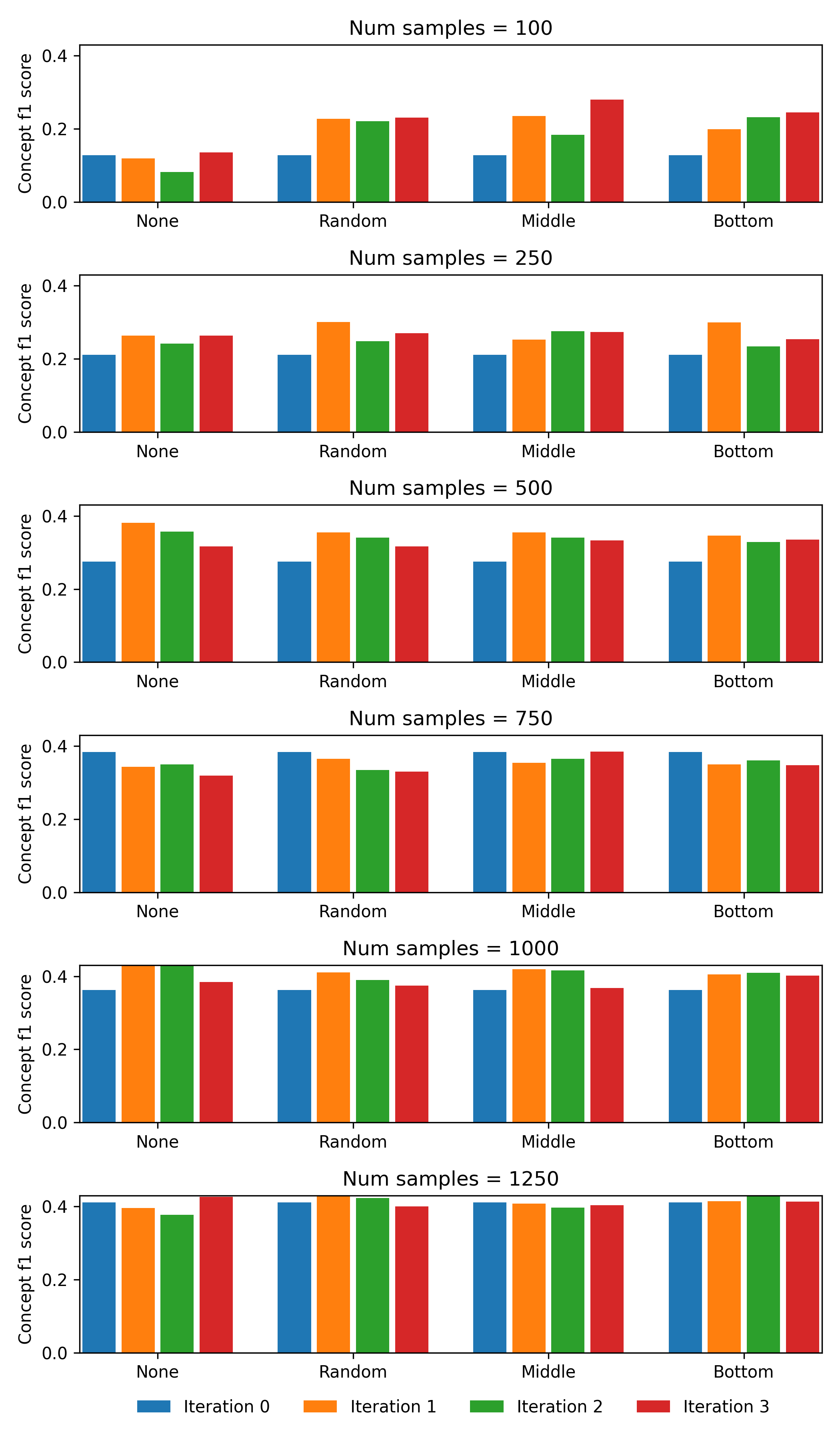}
    \caption{Concept F1 scores for each starting number of examples broken down by human-label sampling strategy and labeling iteration number. Strategy "None" corresponds to no additional human-labeling (only pseudo-labeling). The maximum value of the axis corresponds to the value achieved by training on all 6400 human labels.}
    \label{fig:sampling_strat}
\end{figure}

\end{document}